\documentclass[10pt,twocolumn,letterpaper]{article}
\usepackage{3dv}
\usepackage{times}
\usepackage{epsfig}
\usepackage{graphicx}
\usepackage{subcaption}
\usepackage{comment}
\usepackage{adjustbox}
\usepackage{url}
\usepackage{amsmath}
\usepackage{amssymb}
\usepackage{algorithm}
\usepackage{algorithmicx}
\usepackage{comment}

\threedvfinalcopy 

\def\threedvPaperID{115} 

\ifthreedvfinal\fi
\begin{document}

\title{Multiview RGB-D Dataset for Object Instance Detection}

\author{Georgios Georgakis, Md Alimoor Reza, Arsalan Mousavian, Phi-Hung Le, Jana Ko\v{s}eck{\'a} \\
Department of Computer Science\\
George Mason University\\
{\tt\small \{ggeorgak,mreza,amousavi,ple13,kosecka\}@gmu.edu}
}
\maketitle

\begin{abstract}
This paper presents a new multi-view RGB-D dataset of nine kitchen scenes, each containing several objects in realistic cluttered environments including a subset of objects from the BigBird dataset~\cite{Singh_ICRA14}. The viewpoints of the scenes are densely sampled and objects in the scenes are annotated with bounding boxes and in the 3D point cloud. Also, an approach for detection and recognition is presented, which is comprised of two parts: i) a new multi-view 3D proposal generation method and ii) the development of several recognition baselines using AlexNet~\cite{Krizhevsky_NIPS12} to score our proposals, which is trained either on crops of the dataset or on synthetically composited training images. Finally, we compare the performance of the object proposals and a detection baseline to the Washington RGB-D Scenes (WRGB-D) dataset~\cite{Lai_ICRA14} and demonstrate that our Kitchen scenes dataset is more challenging for object detection and recognition. The dataset is available at: \url{http://cs.gmu.edu/~robot/gmu-kitchens.html}. 


\end{abstract}

\section{Introduction}
The problem of detection and recognition of common household objects in realistic environments is one of the key enabling factors for service robotics. In this paper we present an approach for detection and recognition of object instances in cluttered kitchen scenes, whose scale and size affords manipulation by commonly used robotic grippers or hands. One of the driving forces of progress in detection and categorization is the use of machine learning techniques and subsequent performance evaluation of the proposed approaches on different datasets.

The benchmarks available in computer vision community are largely comprised of images from photo sharing sites with varying amount of clutter~\cite{Lin_ECCV14,Everingham_IJCV10}. With the inception of Microsoft Kinect sensor, several RGB-D datasets also have been released such as the NYUD-V2~\cite{Silberman_ECCV12}, the Washington RGB-D (WRGB-D) scenes dataset~\cite{Lai_ICRA14}, and the BigBird~\cite{Singh_ICRA14}. Each of these datasets has been collected for solving a specific task in mind such as categorization, pose estimation, or object segmentation.
NYUD-V2 offers a large set of objects in a diverse number of scenes but lacks the multi-view aspect and puts less emphasis on the small hand-held objects. On the other hand, the WRGB-D scenes dataset focuses on the small objects, but is limited when it comes to the number of objects and the level of clutter in the scenes. We introduce a new dataset that addresses these shortcomings, by using objects from the BigBird and creating scenes that are more realistic in terms of number of objects available, clutter, and viewpoint variation, while our focus is on the detection of hand-held objects.

\begin{figure}[t]
	\centering
    \includegraphics[width=0.23\textwidth]{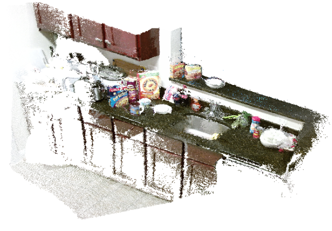}
    ~
    \includegraphics[width=0.23\textwidth]{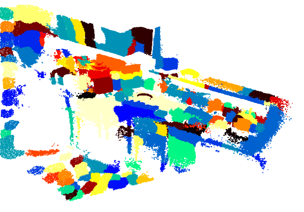}
    \\
    \vspace*{1mm}
    
    \includegraphics[width=0.23\textwidth]{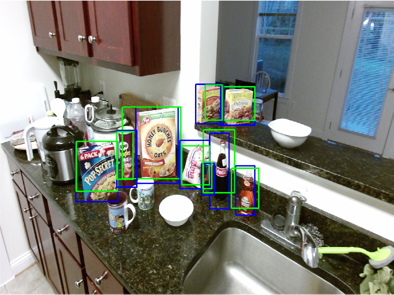}
    ~
    \includegraphics[width=0.23\textwidth]{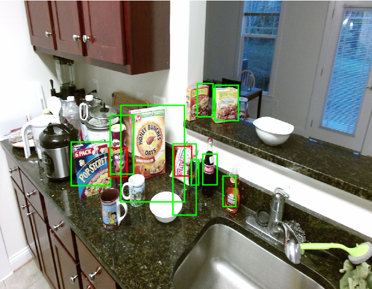}
    
\caption{ Top left: Example reconstructed scene from the Kitchen scenes dataset. Top right: Generation of 3D object proposals using a simple clustering algorithm. Bottom left: Proposals projected on a 2D image in the scene. For clarity only proposals with high overlap with the ground truth are shown. Bottom right: Object recognition using our projected proposals. The blue bounding boxes are the ground truth, the green are correct detections, and the red are false detections.
}
\label{title_img}
\end{figure}

The contributions of the paper are as follows:

i) We present a new RGB-D multi-view kitchen scenes dataset for instance detection and recognition of commonly encountered house-hold objects in realistic settings. The dataset is comprised of densely sampled views of several kitchen counter-top scenes and provides annotation in both 2D and 3D. The complexity of the dataset is demonstrated by comparing to the WRGB-D scenes dataset~\cite{Lai_ICRA14} on the task of object detection.

ii) We develop a multi-view object proposal generation method, which uses only 3D information from the reconstructed scenes. We show comparable results to existing proposal generation methods and demonstrate the effectiveness of the multi-view approach against a 3D single-view approach.


iii) Finally, we utilize our proposals to establish several detection baselines on the Kitchen scenes dataset that includes both Convolutional Neural Network (CNN) based approaches as well as a non CNN-based approach. During training of CNNs, we leverage different training strategies and show how they can affect the performance of the final object detection.

Figure~\ref{title_img} illustrates an overview of the approach.


\section{Related Work}

The problem of object detection has been studied extensively in a variety of domains using image only data or RGB-D images. To position our work, we review few representative approaches that have been applied in similar settings. Traditional methods for object detection in cluttered scenes follow the sliding window based pipelines where efficient methods for feature computation and classifier evaluation were developed such as 
DPM~\cite{Felzenswalb_TPAMI10}. Examples of using these models in the table top settings similar to ours include~\cite{Lai_ICRA11, Song_ECCV12}. 
Another commonly and effectively used strategy for object detection exploited the use of local features 
and correspondences between model reference image and the scene. Object detection and recognition systems that deal with textured household objects such as Collet et al.~\cite{collet_IJRR11} and Tang et al.~\cite{Tang_ICRA12} take advantage of the discriminative nature of the local descriptors.
A disadvantage of these local descriptors is that they usually perform poorly in the presence of non-textured objects, which led to alternative representations that capture object's shape properties such as the Shape Context~\cite{Wang_ACCV07}.

In an attempt to reduce the search space of the traditional sliding window techniques several recent works have concentrated in generating category-independent object proposals. Some representative works include Edge boxes~\cite{Zitnick_ECCV14}, BING~\cite{Cheng_CVPR14}, and Selective search~\cite{Uijlings_IJCV13}. In the RGB-D settings, Mishra et al.~\cite{Mishra_ICRA12} uses object boundaries to guide the detection of fixation points that denote the presence of objects, while Karpathy et al.~\cite{Karpathy_ICRA13} performs object discovery by ranking 3D mesh segments based on objectness scores. Our 3D multi-view approach eliminates large planar surfaces in the scenes to facilitate the segmentation of the small objects. Recently, proposal generation methods based on convolutional neural networks (CNN) have been introduced such as the Multibox~\cite{szegedy_arxiv2014}, the DeepMask~\cite{pinheiro_nips2015}, and the Region Proposal Network (RPN) in~\cite{Ren_NIPS15}. These methods perform very well on the settings they were trained for, but they require re-training
in order to generalize to new settings.

Since the advent of deep learning methods, the choice of features for particular recognition tasks has been replaced by various alternatives for training or fine tuning deep CNNs or design of new architectures and optimization functions suited for various tasks. Early adoption of these techniques using CNNs such as R-CNN~\cite{Girshick_CVPR14} for object detection uses object proposal methods~\cite{Uijlings_IJCV13} to find promising bounding boxes, extract the features using the network of~\cite{Krizhevsky_NIPS12} and trains SVM classifiers to classify each bounding box to different categories. Recently, methods such as YOLO~\cite{Redmon_CVPR16}, SSD~\cite{Liu_ECCV16}, and Faster RCNN~\cite{Ren_NIPS15} drop the use of unsupervised proposal generation techniques and train their networks end-to-end to predict bounding boxes in addition to the classification score for each object category. Although these methods perform significantly well on challenging object detection benchmarks, they require large amounts of bounding box labeled training data.

In the RGB-D table-top settings, the availability of training sets with labeled object instances is limited. In addition, existing datasets are either captured in non realistic settings (Willow Garage~\cite{aldoma_12}), do not focus on the small objects (NYU-V2~\cite{Silberman_ECCV12}, SUN RGB-D~\cite{song_cvpr2015}), or they do not provide a large set of objects in clutter (WRGB-D Scenes~\cite{Lai_ICRA14}). Our kitchen scenes dataset addresses these shortcomings by focusing on the small hand held objects, and by increasing the clutter and viewpoint variation in the scenes. More comprehensive discussion about additional RGB-D datasets can be found in~\cite{Firman_CVPRW16}. Recent works such as Held et al.~\cite{Held_ICRA16} have attempted to address the limitation of training data in these settings by proposing the strategy of pre-training AlexNet~\cite{Krizhevsky_NIPS12} on an auxiliary dataset, before training on a single image and performing object instance recognition. The problem considered, however, was the one of object recognition as opposed to detection in cluttered scenes.

In the following sections we describe the ingredients of our paper starting with a brief discussion of our Kitchen scenes dataset. Then we describe the 3D multi-view proposal generation strategy and its evaluation against 3D single-view and 2D proposals. Finally, we describe the object detection method leveraging the 3D multi-view proposals and establish several object detection baselines on the new Kitchen scene dataset.

\newcommand*\rot{\rotatebox{90}}

\begin{table}[t]
\begin{center}
	\begin{tabular}{|p{3.0em}|p{3.0em}|p{3.0em}|p{4.0em}|}
	\hline
    Scene & Objects & Frames & 3D points \\
    \hline
    1 & 16 & 880 & 13.5x$10^{6}$ \\
    \hline
    2 & 14 & 728 & 13.7x$10^{6}$ \\
    \hline    
    3 & 12 & 763 & 9.7x$10^{6}$\\
    \hline
    4 & 12 & 714 & 5.7x$10^{6}$ \\
    \hline
    5 & 15 & 1316 & 11.6x$10^{6}$ \\
    \hline    
    6 & 15 & 451 & 7.7x$10^{6}$ \\
    \hline
    7 & 11 & 740 & 13.2x$10^{6}$ \\
    \hline
    8 & 10 & 398 & 13.5x$10^{6}$ \\
    \hline    
    9 & 13 & 745 & 8.3x$10^{6}$\\
    \hline    
	\end{tabular}
    \caption{Kitchen scenes dataset statistics.}
    \label{tab:dataset_stats}
\end{center}
\end{table}

\begin{figure*}[t] 
      \centering
      \includegraphics[width=0.3\textwidth]{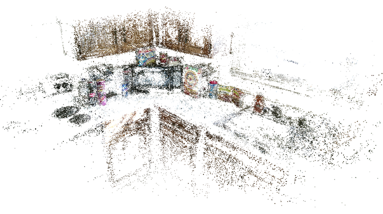}
      \quad
      \includegraphics[width=0.3\textwidth]{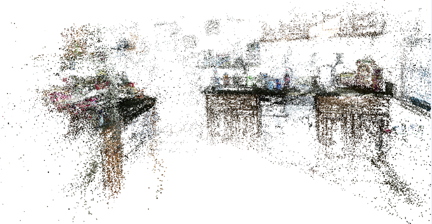}
      \quad
      \includegraphics[width=0.3\textwidth]{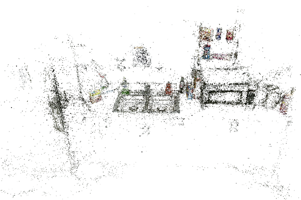}
      \\
      \vspace{1mm}
	  \includegraphics[width=0.3\textwidth]{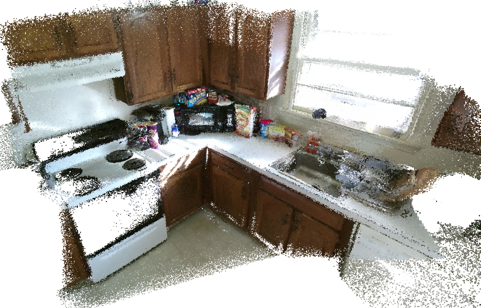}
	  \quad
	  \includegraphics[width=0.3\textwidth]{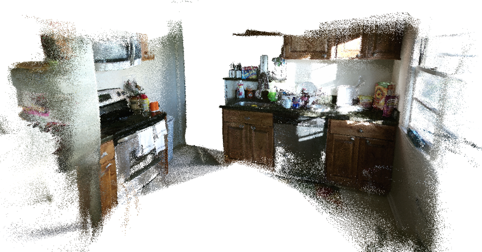}
      \quad
      \includegraphics[width=0.3\textwidth]{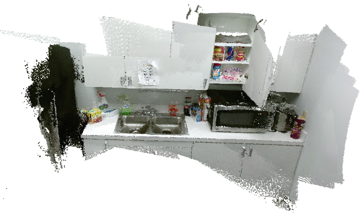}
      \caption{Top row: Sparse reconstructions of our scenes using the state-of-the-art incremental Structure from Motion (SfM) algorithm~\cite{Schoenberger_CVPR16}. Bottom row: Corresponding dense point clouds following the procedure in Section~\ref{sec:dataset}.}
      \label{dense_reconstruction}
\end{figure*}

\section{Kitchen Scenes Dataset}
\label{sec:dataset}
Our Kitchen scenes dataset contains 9 videos from realistic kitchen environments. Table~\ref{tab:dataset_stats} shows scene statistics in terms of \emph{number of objects}, \emph{number of frames}, and \emph{number of points in the dense point cloud}. We focus on small everyday objects which afford manipulation with the commonly used robotic hands or grippers. Each scene contains objects taken from the Big Bird Instance Dataset~\cite{Singh_ICRA14} along with other commonly found objects such as bowls and coffee mugs. The dataset is collected using the Kinect V2 RGB-D sensor of higher resolution ($1920 \times 1080$). It uses time-of-flight camera for depth estimation for more accurate results compared to structured light sensor. The scenes were captured by a person holding the Kinect sensor and moving around in a path, which would allow all objects being seen from a number of different viewpoints and scales. The objects were placed in several different locations, e.g. in shelves, on top of appliances and on kitchen counters with realistic amount of clutter and occlusions. We utilized the open-source Iai Kinect2 library~\cite{Wiedmeyer_UB15} for calibrating the Kinect camera, capturing the data, and aligning the depth channel with RGB image. 


\begin{figure}[t]
	\centering
    \includegraphics[width=0.22\textwidth]{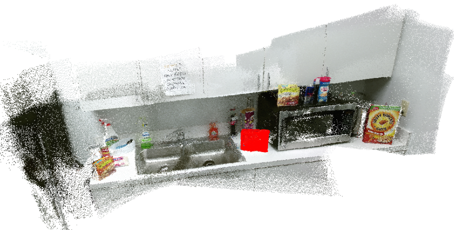}
    \quad
    \includegraphics[width=0.20\textwidth]{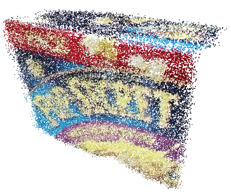}
    \\
    \vspace*{1mm}
    
    \includegraphics[width=0.2\textwidth]{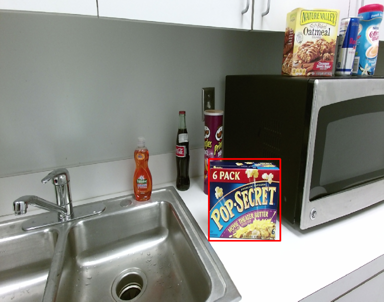}
    \quad
    \includegraphics[width=0.2\textwidth]{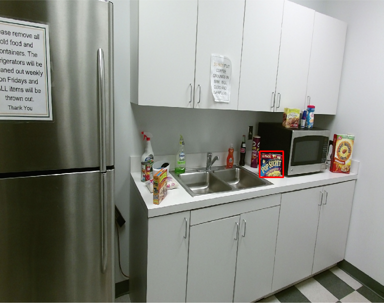}
    
\caption{Annotation procedure for an object in a scene. Top row: Selected 3D points (shown in red in top left) of an object using Meshlab~\cite{Meshlab}. Bottom row: 2D bounding box annotations of the object after projecting the selected points to the images.}
\label{annotation}
\end{figure}
\paragraph{Camera Pose Estimation.}
The video frames are first registered using RGB images utilizing an incremental Structure-from-Motion (SfM) software COLMAP~\cite{Schoenberger_CVPR16}. COLMAP takes a set of images and produces camera poses $C =\{C_{i} \in SE(3) \: | \: i = 1,2,...,N \}$ for registered images along with the sparse reconstruction of the scene as a set of points $P = \{P_{j} \in {\cal R}^3 \: | \: j = 1,2,...,M\}$, where each $P_{i}$ is of the form $P_{i}=[X_{i},Y_{i},Z_{i}]$. This registration process yields sparse reconstruction of the scene up to a universal scale. To determine the scale parameter and to obtain the dense 3D point-cloud, we use the depth channel associated with the RGB-D frames. 
The camera poses from COLMAP are used to project 3D points from each RGB-D frame to the reference  coordinate frame. The 3D point cloud of the depth channel is of the form $p_{i}^d=[x_{i},y_{i},z_{i}]$. To determine the scale between $Z_{i}$ and $z_{i}$ values before projecting into the world coordinate, we use the correspondences between $P_i$'s and $p_i$'s. 
For reliable scale estimation we only consider 3D points whose $Z_{i}$ values are between the 10$^{th}$ and 90$^{th}$ percentile. We then find the ratios between the $Z_{i}$ and $z_{i}$ values and keep their median as the representative scale factor $\alpha$. The scale factor is then used to register the depth maps to the common reference frame. Figure~\ref{dense_reconstruction} shows examples of sparse and dense point clouds of our scenes.




\begin{figure*}[t!] 
      \centering
      \begin{subfigure} [t!] {0.45\textwidth}
      \includegraphics[width=\textwidth]{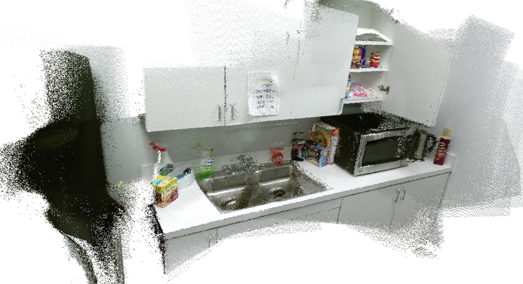}
      \caption{}
      \label{subfig:proposal_scene}
      \end{subfigure}
      ~
      \begin{subfigure} [t!] {0.45\textwidth}
      \includegraphics[width=\textwidth]{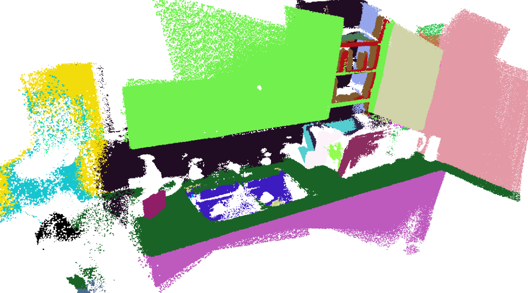}
      \caption{}
      \label{subfig:proposal_planes}
      \end{subfigure}
      ~
      \begin{subfigure} [t!] {0.40\textwidth}
      \includegraphics[width=\textwidth]{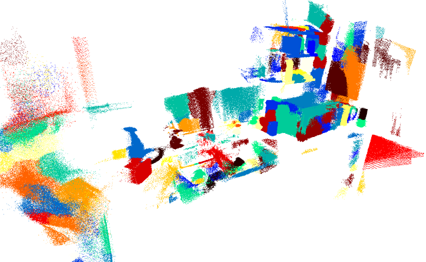}
      \caption{}
      \label{subfig:proposal_clusters}
      \end{subfigure}
      ~
      \begin{subfigure} [t!] {0.45\textwidth}
      \includegraphics[width=\textwidth]{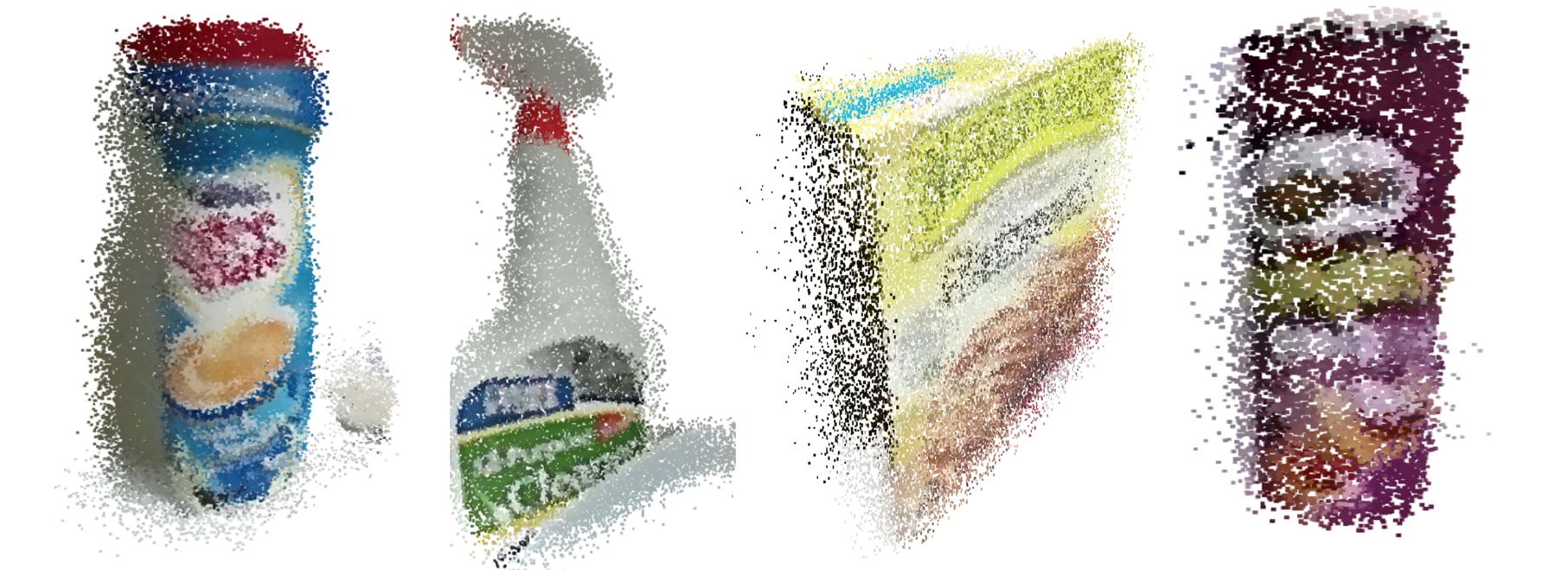}
      \caption{}
      \label{subfig:proposal_ex}
      \end{subfigure}
      
      \caption{3D Object proposal generation: (a) Dense point cloud of a scene; (b) output of the plane detection algorithm; (c) output of the Mean Shift clustering given $radius=0.4$ after removing 33\% of the large planes; (d) extracted 3D proposals after cuboid fitting.}
\label{fig:proposals_3d}
\end{figure*}

\paragraph{Scene Annotation.} We pursue the strategy from~\cite{Lai_ICRA11} for annotating the objects in the scenes. We utilize a feature in the publicly available MeshLab~\cite{Meshlab} tool in order to select groups of points in the registered 3D point cloud. Each group corresponds to a 3D segmentation of an object. We then crop these segmentations from the scenes and project them on the frames in which they are visible. The projected points are then used to define a bounding box over the object. 
Following this strategy, we are able to obtain both 3D segmentation level annotations along with 2D bounding boxes. Figure~\ref{annotation} shows the annotation procedure for an object. 

\section{3D Multi-view Proposal Generation}
\label{sec:proposals}
The object detection strategy we adopt relies on generation of image proposals followed by a feature extraction and classification. We follow a simple algorithm to generate 3D object proposals from the dense point clouds of the scenes and demonstrate that they are superior to object proposals generated from a single RGB-D frame, and comparable to 2D image based RGB proposal generation methods. However, 3D multi-view object proposals can be potentially more useful in tasks such as pose estimation for robotic grasping, and can be more reliable in scenes when objects are occluded. 
Our algorithm starts with the removal of large planar surfaces, followed by mean shift clustering~\cite{Comaniciu_TPAMI02} of the remaining points, and finally removal of outlier points with a cuboid fitting strategy. Figure~\ref{fig:proposals_3d} highlights the procedure.

In order to generate proposals for our objects of interest in 3D, we first remove large supporting surfaces from the scene. This step removes a lot of the unwanted areas of the scene and in some cases it causes object regions to be discontinued in 3D space. A 3D Hough transform based plane detection algorithm is utilized to detect all planar surfaces in the 3D scene~\cite{Borrmann_J3DR11}. Figure~\ref{subfig:proposal_planes} shows our detected planes of a 3D dense point cloud of a scene. The detected planes are sorted by their size and a certain percentage of them is removed. The exact number of planes that needs to be removed vary across scenes. Hence a number of different point clouds of a scene, each with a different number of planes removed, are generated. Specifically, we remove 50\%, 33\%, 25\%, 15\%, and 10\% of the planes in order to create five new 3D point clouds. 
The plane fitting and filtering is followed by mean shift clustering~\cite{Comaniciu_TPAMI02} on each new point cloud. Since mean shift is a density based algorithm it is more applicable to our setting than other approaches as objects in 3D tend to have compact representations. Mean shift clustering requires only one radius parameter. The clustering is applied several times, each time with a different value (ranging from 0.3 to 1.0) for the radius in order to capture objects of various scales. 

The final step of our algorithm is to remove any outlier points in the resulting clusters to get a compact 3D proposals. Similar to~\cite{Lin_ICCV13}, we use direct search, and achieve this by finding the tightest 3D cuboid for each cluster that includes at least 90\% of the points. We relax the problem by not considering any rotation for the cuboid, and use just six degrees of freedom, three for shifting the origin and three for scaling the extent of the cuboid along each axis. The best cuboid is chosen based on the volume to number of included points ratio. The 3D points inside the best-fit cuboid comprise the object proposal.

\begin{figure}[t] 
      \centering
      \includegraphics[width=0.4\textwidth]{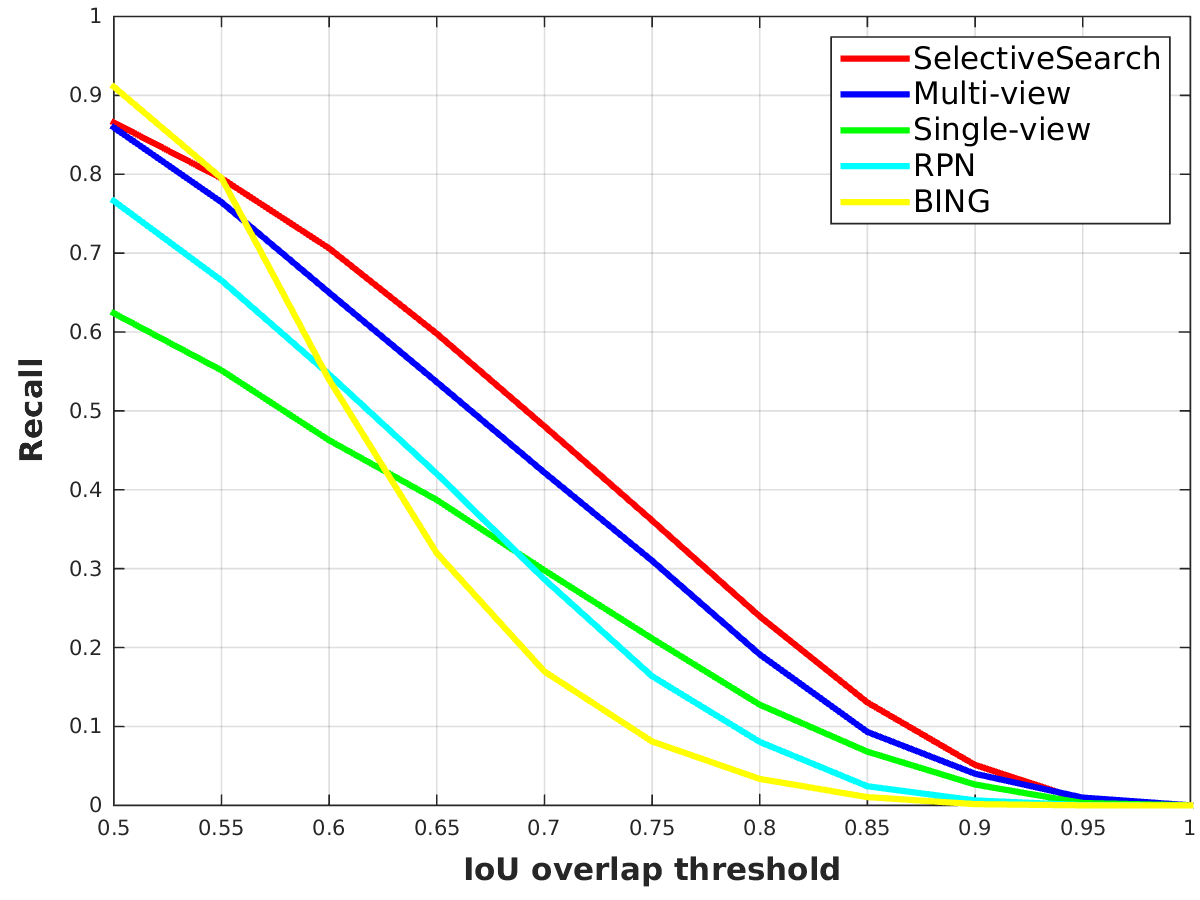}
      \caption{Recall vs IoU overlap threshold for the proposal strategies on the kitchen scenes dataset.}
      \label{fig:iou}
\end{figure}

\paragraph{Evaluation.} The 3D proposals are evaluated on the images since the vast majority of object detection algorithms operate in image space. We compare our multi-view 3D proposals against a single-view 3D proposal generation algorithm, which computes object proposals using 3D point clouds from a single RGB-D frame rather than the dense point cloud of the scene. For a fair comparison, the same pipeline is followed for the generation of the single-view proposals.
We also compare our 3D proposals against two widely used proposal generation algorithms, Selective search~\cite{Uijlings_IJCV13} and BING~\cite{Cheng_CVPR14}, and a CNN approach, the Faster R-CNN RPN~\cite{Ren_NIPS15}. Table~\ref{tab:proposal_results} presents the results in terms of \emph{recall} for an IoU overlap threshold of 0.5, while figure~\ref{fig:iou} illustrates \emph{recall} given a range for the IoU overlap threshold. For all approaches, we have generated around 3000 proposals per image. The performance is reported on the 11 Big Bird objects~\cite{Singh_ICRA14} that have been included in our scenes. Table~\ref{tab:proposal_results} reports also the average performance of the algorithms when the \emph{coke bottle} object is not taken into consideration, since we have observed that our 3D proposals perform poorly for transparent objects 
due to limitations of the Kinect sensor.

\begin{figure}[t]
	\centering
    \includegraphics[width=0.2\textwidth]{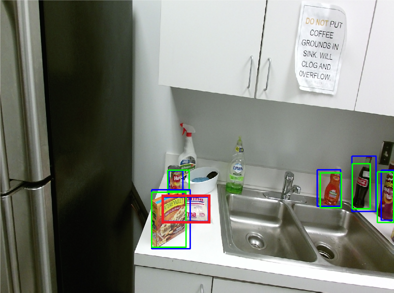}
    ~
    \includegraphics[width=0.2\textwidth]{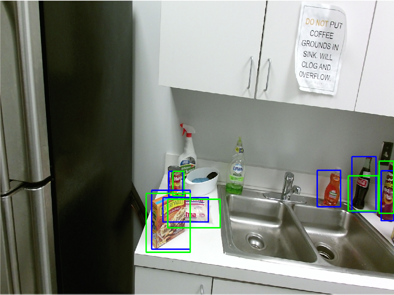}
    \\
    \vspace*{1mm}
    \includegraphics[width=0.2\textwidth]{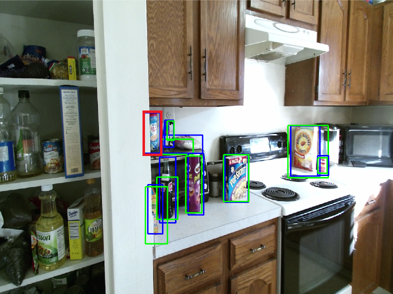}
    ~
    \includegraphics[width=0.2\textwidth]{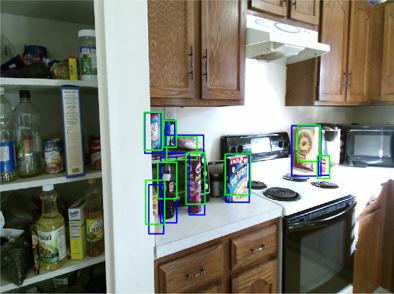}
    \\
    \vspace*{1mm}
    \includegraphics[width=0.2\textwidth]{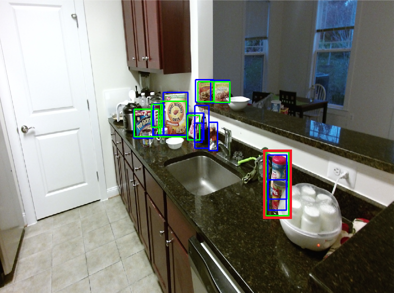}
    ~
    \includegraphics[width=0.2\textwidth]{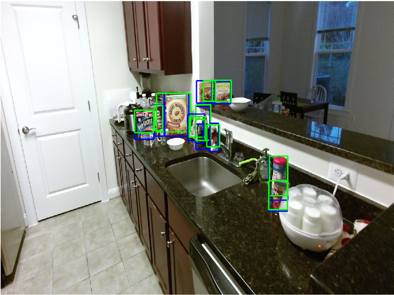}
    \caption{Failure examples from Selective Search~\cite{Uijlings_IJCV13} (left column) where the multi-view is successful (right column). The blue bounding boxes are the ground truth, the green are the proposals with the highest overlap with the ground truth, and the red highlight the failure cases of selective search. The first two rows show missing objects due to occlusions, while in the last row two object are merged in a single proposal.  
 }
    \label{fig:our_ss_comparison}
\end{figure}

The multi-view approach outperforms the single-view by a large margin since the latter cannot recover objects which are heavily occluded or lie on a surface that is not visible from a certain viewpoint. Also, the RPN shows an 11.8\% lower recall than the multi-view approach which signifies that fine-tuning the network might be necessary to achieve comparable results. In comparison to Selective search, multi-view has better recall by a small margin of 1.5\%, when the \emph{coke bottle} is not considered. 
Figure~\ref{fig:our_ss_comparison} shows failure cases of selective search in which the 3D multi-view approach successfully localizes the object. Overall, BING seems to outperform the other strategies, however, its performance drops rapidly while the IoU overlap threshold increases as shown in figure~\ref{fig:iou}. This suggests that BING's proposals have poor overlap with the objects and they may not be suitable to use for classification.

\begin{table*}[t]
\begin{center}
	\begin{tabular}{c|c|c|c|c|c|c|c|c|c|c|c||c||c}
    & \rot{coca cola} & \rot{coffee mate} & \rot{honey bunches} & \rot{hunts sauce} & \rot{mahatma rice} & \rot{nature valley 1} & \rot{nature valley 2} & \rot{palmolive orange} & \rot{pop secret} & \rot{pringles bbq} & \rot{red bull} & \rot{Avg} & \rot{w/o coca cola} \\
    \hline
    Single-view 3D (Ours) & 15.2 & 78.9 & 67.5 & 77.7 & 71.1 & 75.7 & 63.3 & 43.1 & 70.6 & 74.6 & 48.4 & 62.4 & 67.1 \\
    \hline
    Faster R-CNN RPN~\cite{Ren_NIPS15} & 43.6 & 81.4 & 84.1 & 72.7 & 83.4 & 93.8 & 80.9 & \textbf{85.5} & 95.3 & 68.4 & 53.6 & 76.6 & 79.9 \\
    \hline
    Selective Search~\cite{Uijlings_IJCV13} & 59.1 & 95.7 & 94.9 & \textbf{86.5} & \textbf{90.8} & 91.9 & 93.0 & 78.6 & \textbf{96.3} & 90.2 & 75.4 & 86.6 & 89.3 \\
    \hline
    Multi-view 3D (Ours) & 35.8 & \textbf{97.2} & 95.7 & 74.3 & 90.5 & \textbf{97.3} & \textbf{99.2} & 79.2 & 95.6 & 89.9 & 89.0 & 85.8 & 90.8 \\
    \hline
    BING~\cite{Cheng_CVPR14} & \textbf{84.9} & 96.7 & \textbf{97.9} & 83.9 & 86.3 & 92.9 & 92.1 & 81.3 & \textbf{96.3} & \textbf{96.9} & \textbf{92.7} & \textbf{91.1} & \textbf{91.7} \\
    \hline    
	\end{tabular}
    \caption{Recall (\%) results for the proposal strategies on the kitchens scenes dataset.}
    \label{tab:proposal_results}
\end{center}
\end{table*}

We also evaluate a proposal generation method on the WRGB-D scenes dataset~\cite{Lai_ICRA14}, which is a widely used for object detection where the focus is on everyday household objects. It includes 14 video scenes where the objects are placed on a single support surface with low amount of clutter. The support surface usually covers a large percentage of each frame and can be easily segmented out of the image. On the other hand, our Kitchen Scenes dataset contains more realistic environments with objects being placed on several support surfaces such as counters, shelves, tables, and microwave ovens. The support surfaces are frequently occluded or partially visible.

\begin{figure}[t]
	\centering
    \includegraphics[width=0.2\textwidth]{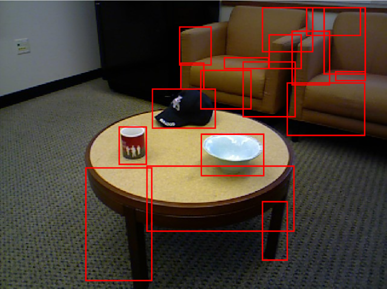}
    ~
    \includegraphics[width=0.2\textwidth]{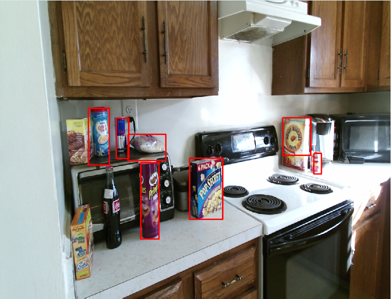}
    \\
    \vspace*{1mm}
    \includegraphics[width=0.2\textwidth]{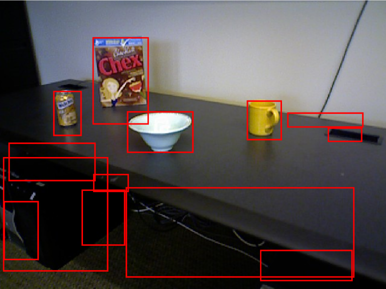}
    ~
    \includegraphics[width=0.2\textwidth]{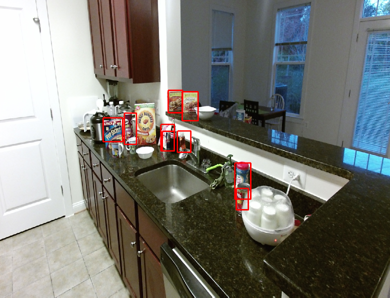}
    \\
    \vspace*{1mm}
    \includegraphics[width=0.2\textwidth]{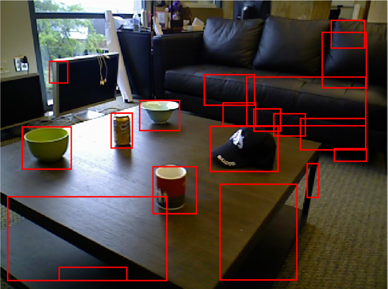}
    ~
    \includegraphics[width=0.2\textwidth]{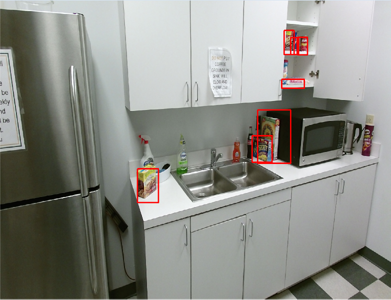}
    \caption{Examples of proposal generation using the single-view algorithm. Left column: Images from the WRGB-D Scenes dataset. Right column: Images from the Kitchen scenes dataset, where only proposals with sufficient overlap with ground truth are shown for clarity.}
    \label{fig:wash_our_proposals}
\end{figure}

To compare with the WRGB-D Scenes dataset, we test the performance of the 3D single-view proposal generation algorithm against its performance on our Kitchen scenes dataset.
As can be seen in Table~\ref{tab:wrgb_comparison}, for the WRGB-D dataset the algorithm achieves a recall result that is significantly higher compared to recall results in our Kitchen scenes dataset, with much less proposals per image. Note that in the case of the WRGB-D, we used just a single radius value when computing the mean shift clustering. This result is not surprising for the WRGB-D case since the algorithm detects the support surface in the vast majority of the frames and the clustering segments the objects successfully due to the low amount of clutter. Figure~\ref{fig:wash_our_proposals} shows examples of proposals generated on images from both datasets.

\begin{table}
\begin{center}
	\begin{tabular}{|c|c|}
    \hline
    Dataset & Recall(\%)\space/\space No. Proposals \\
    \hline
    WRGB-D~\cite{Lai_ICRA14} & 89.3\space/\space17\\
    \hline
    Our Kitchen Scenes & 62.4\space/\space2989\\
    \hline    
    \end{tabular}
    \caption{Recall results given an average number of proposals per frame when applying the single-view proposal generation algorithm from section~\ref{sec:proposals} on the WRGB-D and our Kitchen Scene dataset. The recall was estimated using an IoU threshold of 0.5.}
    \label{tab:wrgb_comparison}
\end{center}
\end{table}

\section{Object Detection}
The object proposals from section~\ref{sec:proposals} are category-agnostic, therefore further processing is needed to determine the object categories. Towards this goal, we establish four baselines for the object detection task on the proposed Kitchen scenes dataset. The baselines differ in the approach or the type of training data used.  

1.{\em Turntable.} We use the turntable images provided by the BigBird dataset~\cite{Singh_ICRA14}, which depict a single object against a clean background for training the CNN.

2.{\em Turntable Background.} 
We use the turntable images provided by the BigBird dataset~\cite{Singh_ICRA14} superimposed on the backgrounds randomly sampled from the kitchen scenes. 

3.{\em CNN Scene Folds.} We use folds of images from the dataset along with bounding boxes of the objects of interest as training examples. 
We follow recent trends in object detection and train a Convolutional Neural Network (CNN) for three of our baselines.

4.{\em HMP Scene Folds.} In addition to the three baselines, we follow the non CNN route of Hierarchical Matching Pursuit (HMP) for the fourth baseline~\cite{Bo_ISER12}.
\\

\noindent\textbf{Turntable.} We use the images with single object placed on a clean background from the BigBird dataset~\cite{Singh_ICRA14}. We sub-sample these images where objects are observed from 60 viewpoints. Five random crops of each image are created in order to increase the variability of the training set. For the background category, we randomly sample patches from kitchen scenes of the NYUD-V2 dataset~\cite{Silberman_ECCV12}. The size of the generated training set is 3600 images. We feed these training images to a Convolutional Neural Network (CNN) for training. We refer to this baseline as \emph{Turntable}.

\begin{figure}[t] 
      \centering
       \includegraphics[width=0.12\textwidth]{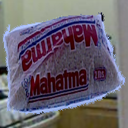}
       ~
       \includegraphics[width=0.12\textwidth]{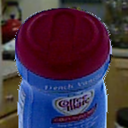} 
        ~
       \includegraphics[width=0.12\textwidth]{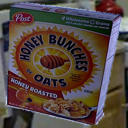} 
      \caption{Training examples from the \emph{Turntable Background} baseline of objects superimposed against random backgrounds from the NYUD-V2~\cite{Silberman_ECCV12} kitchen scenes.}
      \label{fig:bg}
\end{figure}

\noindent\textbf{Turntable Background.} The second baseline extends the training set of \emph{Turntable} by superimposing the object mask image from BigBird dataset on random background patches from NYUD-V2 Kitchen scenes~\cite{Silberman_ECCV12}. These synthesized training images are prepared to make our detector more robust to different backgrounds. Examples of these synthetic patches are shown in Figure~\ref{fig:bg}. Objects that have incomplete ground truth segmentation masks in the Big Bird due to sensor limitations are not superimposed on background patches. We train another CNN with this training set containing 6525 images and refer to this baseline as \emph{Turntable Background}. AlexNet~\cite{Krizhevsky_NIPS12} is used for both of the Turntable and Turntable Background baselines as 
initialization and is trained for 10000 iterations with a learning rate of 0.0001. During testing, the trained models are applied on the proposals generated from the multi-view approach on the scene frames. The motivation for this experiment is to evaluate the object detection performance when realistic annotated data - objects placed in a realistic scene - are not available. 

\noindent\textbf{CNN Scene Folds.}
This experiment investigates the performance of a detector that is trained on examples extracted from our Kitchen scenes images. We do a 3-Fold cross validation experiment using a random partition of the nine scenes into folds. Each training fold contains images from six out of our nine scenes. The images from the remaining three scenes are included in the test fold. 
The 3D multi-view proposals are used to generate examples from the training fold scenes. We follow the approach of R-CNN~\cite{Girshick_CVPR14} and consider all proposals with an IoU overlap larger than 0.5 with the ground truth as instances of a particular object. Proposals having an IoU less than 0.3 are considered as background. We ignore the rest of the proposals having values in between these two ranges. We sub-sample proposals every 10 images to avoid high correlation between the examples and end up with a training set of around 70000 patches for each fold. Following the CNN architecture of \emph{Turntable} and \emph{Turntable Background} baselines, we use the AlexNet for initialization, and train each fold separately for 30000 iterations with a learning rate of 0.0001. 

\noindent\textbf{HMP Scene Folds.}
We also compare our detection against a non-CNN architecture based feature generation method. We selected Hierarchical Matching Pursuit (HMP) approach, which is used in the context of detecting, labeling, and segmenting objects similar to our Kitchen Scenes dataset environment~\cite{Lai_ICRA14,Bo_ISER12}. HMP is a sparse coding technique that learns features in an unsupervised manner from the training images. We applied HMP on our 3-Fold cross-validation experiment using the publicly available implementation provided by the authors~\cite{Bo_ISER12}. In each fold, we learned dictionaries from both the grayscale and RGB channels of the proposal-patches of the training fold. The learned dictionaries are used to extract features, i.e., the sparse codes of the patches, which are then concatenated together. The dictionary sizes in the 1st and 2nd layer are 75 and 150 respectively. We used patch sizes of 5x5 and 4x4 respectively for learning the dictionaries in these two layers. A linear SVM is trained on the features of the training fold. 
\begin{table*}[t]
\begin{center}
	\resizebox{16cm}{!}{ 
	\begin{tabular}{c|c|c|c|c|c|c|c|c|c|c|c|c|c}
    & \rot{coca cola} & \rot{coffee mate} & \rot{honey bunches} & \rot{hunts sauce} & \rot{mahatma rice} & \rot{nature valley 1} & \rot{nature valley 2} & \rot{palmolive orange} & \rot{pop secret} & \rot{pringles bbq} & \rot{red bull} & \rot{Background} & \rot{mAP}\\
    \hline
    Turntable & 1.0 & 25.7 & 8.6 & 2.8 & 17.2 & 21.2 & 24.0 & 7.6 & 40.1 & 1.2 & 2.6 & 89.5 & 20.1 \\
    \hline
    Turntable Background & 0.1 & 33.0 & 15.9 & 17.9 & 18.0 & 19.9 & 26.0 & 10.8 & 32.5 & 3.2 & 3.3 & 89.5 & 22.5\\
    \hline
    \hline
    HMP Scene Folds & 0.0 & 26.8 & 22.8 & 13.2 & 2.7 & 33.7 & 17.2 & 4.1 & 14.3 & 11.5 & 8.0 & 86.5 & 20.1 \\
    \hline    
    CNN Scene Folds & \textbf{3.5} & \textbf{48.8} & \textbf{50.0} & \textbf{27.6} & \textbf{27.9} & \textbf{52.4} & \textbf{48.1} & \textbf{18.8} & \textbf{53.6} & \textbf{46.9} & \textbf{32.7} & \textbf{90.6} & \textbf{41.7} \\
    \hline
	\end{tabular}
    }
    \caption{Average precision (\%) results for the object detection baselines on the kitchen scenes dataset.}
    \label{tab:detection_results}
\end{center}
\end{table*}

\begin{table}[t]
\begin{center}
	\resizebox{8.5cm}{!}{ 
	\begin{tabular}{c|c|c|c|c|c|c|c}
     WRGB-D~\cite{Lai_ICRA14} & \rot{bowl} & \rot{cap} & \rot{cereal box} & \rot{coffee mug} & \rot{soda can} & \rot{background} & \rot{mAP} \\
    \hline
	Turntable & 42.2 & 71.7 & 72.3 & 49.0 & 57.7 & 82.5 & 62.6 \\
    \hline
	\end{tabular}
    }
   	\caption{Object detection results for the WRGB-D Scenes dataset~\cite{Lai_ICRA14}. The detection was performed using the procedure of the Turntable baseline for this dataset.}
    \label{tab:washington_detection_res}
\end{center}
\end{table}
\paragraph{Discussion.} The detection results for all baselines are illustrated in Table~\ref{tab:detection_results}.
CNN Scene Folds outperforms the next best baseline by 19.2\% in terms of mean Average Precision (mAP), while the HMP Scene Folds and Turntable produce the lowest performance.  
It is evident that training on examples which contain similar backgrounds as the test scenes leads to better performing detectors, however, these type of training data are usually harder to acquire. Regarding the Turntable and Turntable Background baselines, we notice that the inclusion of the examples with random backgrounds increases mAP by 2.4\%.
This increase in performance suggests that Turntable Background is more robust to detecting objects in novel backgrounds. However, this increase is small which suggests that more sophisticated approaches than simply randomly choosing backgrounds might be required to achieve significant increase in the performance. Held et al~\cite{Held_ICRA16} achieved similar performance on their experiment on the WRGB-D scenes dataset~\cite{Lai_ICRA11} where they also trained on turntable images superimposed against random backgrounds. However, there are three differences between their experiment and ours. First, they sample backgrounds from the same environments as the test set, as provided by the dataset. Second, they finetune their model initially on the Big Bird dataset, while they only use one training example from the WRGB-D dataset~\cite{Lai_ICRA11} for each object. Finally, they evaluate on the recognition task by cropping  the objects from the scenes using the ground truth annotation, while we utilize our proposal algorithm to perform detection.

Additionally, we investigate the performance of the Turntable baseline on the WRGB-D Scenes dataset~\cite{Lai_ICRA14}. We sampled the turntable images provided in the dataset to create a training set of 20500 examples for the five object categories available in the scenes. We again initialize with the AlexNet and train for 10000 iterations with a learning rate of 0.0001. For the detection we used the single-view proposals which work very well on that dataset (see Table~\ref{tab:wrgb_comparison}). The results are shown in Table~\ref{tab:washington_detection_res}. It is noticeable that the current baseline achieves a high performance in this dataset, while at the same time performs very poorly in the Kitchen scenes dataset.

We demonstrate the qualitative results of the detection baselines in the supplementary materials.

\section{Conclusions}
We have presented a new RGB-D multi-view dataset for object instance detection and recognition of commonly encountered house hold objects in realistic settings. The dataset was utilized to demonstrate the effectiveness of a novel 3D multi-view object proposal method against a 3D single-view method, while at the same time achieving comparable results to established image-based proposal methods. The generated proposals were further used to establish and compare several object detection pipelines on the dataset.
These include both deep-learning-based approaches and a non-deep-learning-based approach. Not surprisingly the performance of the CNN based strategies is superior to previously used methods. Also, the best performance is achieved by training and testing on the different folds from the kitchen dataset. Note that the performance drops significantly when the dataset used for training does not contain the type of backgrounds found in test, e.g. training with random backgrounds.
Comparative experiments on the WRGB-D dataset reveal differences in nature of complexity with respect to the new Kitchens scenes dataset.

\paragraph{Acknowledgments.} We acknowledge support from NSF NRI grant 1527208. Some of the experiments were run on ARGO, a research computing cluster provided by the Office of Research Computing at George Mason University, VA. (URL: http://orc.gmu.edu).


{\small
\bibliographystyle{ieee}
\bibliography{egbib}
}

\end{document}